\documentclass{article}

\usepackage{arxiv}
\usepackage{float} 
\usepackage[utf8]{inputenc} 
\usepackage[T1]{fontenc}    
\usepackage{hyperref}       
\usepackage{url}            
\usepackage{booktabs}       
\usepackage{amsfonts}       
\usepackage{nicefrac}       
\usepackage{microtype}      
\usepackage{lipsum}		
\usepackage{graphicx}
\usepackage{natbib}
\usepackage{doi}
\usepackage{tabularx} 
\usepackage{booktabs} 
\usepackage{algorithm}
\usepackage{algpseudocode}
\usepackage{amsmath}
\usepackage{adjustbox}

\title{Demonstrating the Efficacy of Kolmogorov-Arnold Networks in Vision Tasks}

\author{
    Minjong Cheon \\
    \texttt{jmj541826@gmail.com} \\
}

\date{Independent Researcher}

\renewcommand{\shorttitle}{\textit{arXiv} Template}

\hypersetup{
pdftitle={A template for the arxiv style},
pdfsubject={q-bio.NC, q-bio.QM},
pdfauthor={Minjong},
pdfkeywords={First keyword, Second keyword, More},
}

\begin{document}
\maketitle

\begin{abstract}
In the realm of deep learning, the Kolmogorov-Arnold Network (KAN) has emerged as a potential alternative to multilayer projections (MLPs). However, its applicability to vision tasks has not been extensively validated. In our study, we demonstrated the effectiveness of KAN by using KAN-Mixer for vision tasks through multiple trials on the MNIST, CIFAR10, and CIFAR100 datasets, using a training batch size of 32. Our results showed that while KAN outperformed the original MLP-Mixer on CIFAR10 and CIFAR100, it performed slightly worse than the state-of-the-art ResNet-18. These findings suggest that KAN holds significant promise for vision tasks, and further modifications could enhance its performance in future evaluations. Our contributions are threefold: first, we showcase the efficiency of KAN-based algorithms for visual tasks; second, we provide extensive empirical assessments across various vision benchmarks, comparing KAN's performance with MLP-Mixer, CNNs, and Vision Transformers (ViT); and third, we pioneer the use of natural KAN layers in visual tasks, addressing a gap in previous research. This paper lays the foundation for future studies on KANs, highlighting their potential as a reliable alternative for image classification tasks. \end{abstract}

\keywords{Kolmogorov-Arnold Network \and KAN-Mixer \and Vision Tasks \and Image Classification \and Deep Learning}

\section{Introduction}
The Kolmogorov-Arnold Network (KAN) has been presented in the field of deep learning and has the potential to replace multilayer projections (MLPs). The Kolmogorov-Arnold Network (KAN) has been presented in the field of deep learning and has the potential to replace multilayer projections (MLPs). By combining the advantages of splines and MLPs, KAN) performs better than conventional Multi-Layer Perceptrons (MLPs) and consequently lessens the drawbacks of each. KANs increase accuracy in a wide range of tasks by employing the Kolmogorov-Arnold Representation Theorem, which allows them can effectively represent complicated multivariate functions. They are more computationally efficient than MLPs because of a unique architecture that substitutes learnable activation functions for linear weight matrices. Improved interpretability is one of the most compelling advantages of KANs as they simplify complicated functions into simpler univariate components, making it possible to interpret model predictions more clearly \cite{ismayilova2024kolmogorov} \cite{liu2024kan} \cite{cheon2024kolmogorov}. 

This article aims to prove KAN’s performance in vision tasks, while most prior efforts have mostly replaced MLPs in architectures such as CNNs or U-Nets or applied KANs to tabular datasets \cite{kiamari2024gkan}, \cite{xu2024kolmogorovarnold}, \cite{xu2024fourierkangcf}, \cite{bozorgasl2024wavkan}, \cite{abueidda2024deepokan}, \cite{genet2024tkan}, \cite{azam2024suitability}, \cite{li2024ukan}. Instead of including additional algorithms like CNNs or ViTs, we chose KAN-Mixer which solely utilizes KAN layers.  This architecture shares several similarities with the MLP-Mixer, including operating directly on patches and maintaining an equal resolution and size representation throughout all levels. Through specialized modules, the KAN-Mixer executes channel mixing as well as spatial mixing. The Channel-mixing KANs function independently on each channel to enable communication over several channels inside a single token \cite{tolstikhin2021mlp}.

\begin{enumerate}
    \item Introduction of a Novel Approach: This paper introduces a KAN-based approach to vision tasks utilizing only KAN layers, which is KAN-Mixer. Furthermore, we tried several experiments to figure out the effects of the hyper parmeters in the model.
    
    \item Empirical Evaluation: We conducted a comprehensive empirical evaluation of the KAN-Mixer across various vision benchmarks, comparing its performance against MLP-Mixer, CNNs, and Vision Transformers (ViTs). This evaluation highlights the strengths and potential advantages of the KAN-Mixer architecture.
    
    \item Pioneering Performance in Vision Tasks: As existing studies have not extensively demonstrated the performance of KANs in vision tasks, our approach pioneers the application of KANs in this domain.
    
\end{enumerate}

\section{Materials and Methods}

\subsection{Kolmogorov-Arnold Networks}

The Kolmogorov-Arnold representation theorem serves as the inspiration for KAN, a sophisticated form of neural network. Learnable activation functions are present on edges of KANs, in contrast to fixed activations at nodes in classic MLPs. B-splines, piecewise polynomial functions defined by control points and knots, parameterize these activation functions. In KANs, spline-parameterized functions \( \varphi_{q,p} \) convert each input feature \( x_p \), aggregate into intermediate values for each \( q \), and then process the intermediate values using functions \( \Phi_q \). The total of these modified values is the final output \( f(\mathbf{x}) \) which enables KANs to effectively capture intricate patterns in data. The activation functions in KANs integrate a spline component \( \text{spline}(x) = \sum_i c_i B_i(x) \), where \( B_i(x) \) are B-spline basis functions and \( c_i \) are learned coefficients during training, with a Basis Function, normally the Sigmoid Linear Unit (SiLU), defined as \( \text{silu}(x) = \frac{x}{1 + e^{-x}} \) \cite{liu2024kan}.

The formula for the Kolmogorov-Arnold Network (KAN) is:

\[
f(\mathbf{x}) = \sum_{q=1}^{2n+1} \Phi_q \left( \sum_{p=1}^{n} \varphi_{q,p}(x_p) \right)
\]

Here, \( \varphi_{q,p}(x_p) \) are the spline functions, and \( \Phi_q \) are the transformations.

The activation function \( \varphi(x) \) is defined as:

\[
\varphi(x) = w \left( b(x) + \text{spline}(x) \right)
\]

Where \( w \) is a weight, \( b(x) \) is the basis function (implemented as \( \text{silu}(x) \)), and \( \text{spline}(x) \) is the spline function.

The basis function \( b(x) \) is:

\[
b(x) = \text{silu}(x) = \frac{x}{1 + e^{-x}}
\]

The spline function \( \text{spline}(x) \) is:

\[
\text{spline}(x) = \sum_i c_i B_i(x)
\]

\subsection{KAN-Mixer Architecture}

\subsubsection{Input Image}
The model begins with an input image $\mathbf{X}$ of shape $[B, C, H, W]$, where $B$ is the batch size, $C$ is the number of channels, $H$ is the height, and $W$ is the width of the image. This image serves as the raw data that will be processed by the model.

\subsubsection{Image to Patches}
The input image $\mathbf{X}$ is divided into non-overlapping patches using the \textit{ImageToPatches} module. Each patch is of size $P \times P$, resulting in a total of $\left(\frac{H}{P} \times \frac{W}{P}\right)$ patches. The image is reshaped into a sequence of patches $\mathbf{X}_{patches}$ with shape $[B, N, P^2 \cdot C]$, where $N = \left(\frac{H}{P} \times \frac{W}{P}\right)$ is the number of patches, and $P^2 \cdot C$ is the number of pixels per patch.

Mathematically, this can be expressed as:
\[
\mathbf{X}_{patches} = \text{reshape}(\mathbf{X}, [B, N, P^2 \cdot C])
\]

\subsubsection{Per-Patch KAN}
Each patch in $\mathbf{X}_{patches}$ is independently processed by the \textit{PerPatchKAN} module, which applies a KANLinear transformation to project each patch into a higher-dimensional space. This transformation enhances the representation of the patch data.

Let $\mathbf{X}_p$ be a patch from $\mathbf{X}_{patches}$. The KANLinear transformation can be represented as:
\[
\mathbf{X}_p' = \text{KANLinear}(\mathbf{X}_p)
\]
where $\mathbf{X}_p'$ is the transformed patch.

\subsubsection{Mixer Stack}
The \textit{MixerStack} consists of multiple layers, each containing two main types of KANLinear-based MLPs: Token Mixing KAN and Channel Mixing KAN.

\paragraph{Token Mixing KAN}
This layer allows communication between different spatial locations (tokens). Each token is independently transformed by KANLinear layers that mix spatial information while keeping channel information constant. For a token $\mathbf{T}$:
\[
\mathbf{T}' = \text{KANLinear}_{token}(\mathbf{T})
\]

\paragraph{Channel Mixing KAN}
This layer allows communication between different channels within each token. Each channel is independently transformed by KANLinear layers that mix channel information while keeping spatial information constant. For a channel $\mathbf{C}$:
\[
\mathbf{C}' = \text{KANLinear}_{channel}(\mathbf{C})
\]

Each layer in the MixerStack alternates between token mixing and channel mixing, effectively capturing both spatial and channel dependencies.

\subsection{Output KAN}
The final \textit{OutputKAN} module aggregates the information from the transformed patches. It first applies a layer normalization to stabilize the feature representation:
\[
\mathbf{X}_{norm} = \text{LayerNorm}(\mathbf{X})
\]

Next, it computes the mean across the token dimension to obtain a fixed-size representation:
\[
\mathbf{X}_{mean} = \text{mean}(\mathbf{X}_{norm}, \text{dim}=1)
\]

Finally, a KANLinear layer projects the aggregated representation to the desired output dimension:
\[
\mathbf{Y} = \text{KANLinear}_{output}(\mathbf{X}_{mean})
\]

\subsection{Final Output}
The final output $\mathbf{Y}$ is produced, which has the shape $[B, n_{output}]$, where $n_{output}$ is the number of output classes or the desired dimensionality of the output.

\begin{algorithm}
\caption{KAN-Mixer Forward Pass}\label{alg:kan-mixer}
\begin{algorithmic}[1]
\Require $\mathbf{X}$: Input image of shape $[B, C, H, W]$
\Ensure $\mathbf{Y}$: Output of shape $[B, n_{output}]$
\State $\mathbf{X}_{patches} \gets \text{reshape}(\mathbf{X}, [B, N, P^2 \cdot C])$ \Comment{Image to Patches}
\ForAll {patch $\mathbf{X}_p \in \mathbf{X}_{patches}$}
    \State $\mathbf{X}_p' \gets \text{KANLinear}(\mathbf{X}_p)$ \Comment{Per-Patch MLP}
\EndFor
\State $\mathbf{X}_{patches} \gets \{\mathbf{X}_p'\}$ \Comment{Update patches}
\For {layer $l \in \{1, \ldots, L\}$}
    \ForAll {token $\mathbf{T} \in \mathbf{X}_{patches}$}
        \State $\mathbf{T}' \gets \text{KANLinear}_{token}(\mathbf{T})$ \Comment{Token Mixing MLP}
    \EndFor
    \State $\mathbf{X}_{tokens} \gets \{\mathbf{T}'\}$ \Comment{Update tokens}
    \ForAll {token $\mathbf{T} \in \mathbf{X}_{tokens}$}
        \ForAll {channel $\mathbf{C} \in \mathbf{T}$}
            \State $\mathbf{C}' \gets \text{KANLinear}_{channel}(\mathbf{C})$ \Comment{Channel Mixing MLP}
        \EndFor
    \EndFor
    \State $\mathbf{X}_{patches} \gets \{\mathbf{C}'\}$ \Comment{Update channels}
\EndFor
\State $\mathbf{X}_{norm} \gets \text{LayerNorm}(\mathbf{X}_{patches})$ \Comment{Normalize}
\State $\mathbf{X}_{mean} \gets \text{mean}(\mathbf{X}_{norm}, \text{dim}=1)$ \Comment{Aggregate}
\State $\mathbf{Y} \gets \text{KANLinear}_{output}(\mathbf{X}_{mean})$ \Comment{Output MLP}
\State \Return $\mathbf{Y}$
\end{algorithmic}
\end{algorithm}

\section{Result}
Three typical datasets—MNIST, CIFAR-10, and CIFAR-100—were used in our empirical assessment of the KAN-Mixer architecture. The selection of these datasets was based on their extensive usage and recognition in the computer vision field, enabling a reliable comparison with current architectures like MLP-Mixer, CNNs, and VITs. In this study, we conducted experiments to identify the optimal values for two critical parameters in our model's KAN layer: \texttt{n\_channels} and \texttt{n\_hiddens}. These parameters play a significant role in the performance and efficiency of the neural network. To determine the best configuration, we systematically varied the values of \texttt{n\_channels} and \texttt{n\_hiddens} and evaluated the performance of the model on multiple datasets, including CIFAR10, CIFAR100, and MNIST. We measured key performance metrics such as epoch time, test time, GPU memory usage, and test accuracy. For the experiment, we used L4 GPU from Google Colab.

Through this extensive experimentation, we found that setting \texttt{n\_channels} to 64 and \texttt{n\_hiddens} to 128 yielded the best balance of performance and resource utilization across all tested datasets with 10 epochs. The summarized performance metrics for these optimal parameter values are presented in the Tables below. Furthermore, the number of layers did not produce a distinctive performance difference.

\begin{table}[h!]
    \centering
    \small
    \begin{adjustbox}{max width=\textwidth}
    \begin{tabular}{lccccc}
        \toprule
        \textbf{Dataset} & \textbf{Epoch Time (s)} & \textbf{Test Time (s)} & \textbf{GPU Memory Allocated (MB)} & \textbf{GPU Memory Reserved (MB)} & \textbf{Test Accuracy} \\
        \midrule
        CIFAR10 & 42.92 & 5.14 & 26.60 & 420.0 & 0.5924 \\
        CIFAR100 & 43.09 & 5.32 & 26.82 & 433.2 & 0.2489 \\
        MNIST & 44.92 & 5.15 & 29.51 & 432.0 & 0.9730 \\
        \bottomrule
    \end{tabular}
    \end{adjustbox}
    \caption{Summary of Performance Metrics for Various Datasets (n\_channels: 16, n\_hidden: 32)}
    \label{tab:performance_metrics}
\end{table}

\begin{table}[h!]
    \centering
    \small
    \begin{adjustbox}{max width=\textwidth}
    \begin{tabular}{lccccc}
        \toprule
        \textbf{Dataset} & \textbf{Epoch Time (s)} & \textbf{Test Time (s)} & \textbf{GPU Memory Allocated (MB)} & \textbf{GPU Memory Reserved (MB)} & \textbf{Test Accuracy} \\
        \midrule
        CIFAR10 & 43.22 & 5.49 & 55.68 & 846.0 & 0.6300 \\
        CIFAR100 & 43.34 & 5.52 & 56.13 & 846.4 & 0.3190 \\
        MNIST & 46.01 & 5.64 & 61.80 & 852.0 & 0.9732 \\
        \bottomrule
    \end{tabular}
    \end{adjustbox}
    \caption{Summary of Performance Metrics for Various Datasets (n\_channels: 32, n\_hidden: 64)}
    \label{tab:performance_metrics}
\end{table}

\begin{table}[h!]
    \centering
    \small
    \begin{adjustbox}{max width=\textwidth}
    \begin{tabular}{lccccc}
        \toprule
        \textbf{Dataset} & \textbf{Epoch Time (s)} & \textbf{Test Time (s)} & \textbf{GPU Memory Allocated (MB)} & \textbf{GPU Memory Reserved (MB)} & \textbf{Test Accuracy} \\
        \midrule
        CIFAR10 & 110.14 & 11.26 & 77.56 & 1704.0 & 0.6693 \\
        CIFAR100 & 110.16 & 11.22 & 78.44 & 1705.4 & 0.3549 \\
        MNIST & 132.12 & 11.32 & 90.76 & 1692.0 & 0.9816 \\
        \bottomrule
    \end{tabular}
    \end{adjustbox}
    \caption{Summary of Performance Metrics for Various Datasets (n\_channels: 64, n\_hidden: 128)}
    \label{tab:performance_metrics_3}
\end{table}

\begin{table}[h!]
    \centering
    \small
    \begin{adjustbox}{max width=\textwidth}
    \begin{tabular}{lcccc}
        \toprule
        \textbf{Model} & \textbf{GPU Usage (gb)} & \textbf{MNIST} & \textbf{CIFAR10} & \textbf{CIFAR100} \\
        \midrule
        KKAN (Small) & 1.8119 & \textbf{98.90\%} & - & - \\
        Conv \& KAN & - & 98.75\% & - & - \\
        MLP Mixer-5 & 1.5 & - & 60.26 & 34.81 \\
        ViT-10/4 & 14.7 & - & 57.53 & 30.80 \\
        ResNet-18 & 0.6 & - & \textbf86.29 & \textbf59.15 \\
        \textbf{Ours} & \textbf{above table} & 98.16\% & {66.93\%} & {35.49\%} \\
        \bottomrule
    \end{tabular}
    \end{adjustbox}
    \caption{Comparison of Models on MNIST, CIFAR10, and CIFAR100 Datasets}
    \label{tab:comparison}
\end{table}

Using a training batch size of 32, we conducted tests to compare the performance of several models on the MNIST, CIFAR10, and CIFAR100 datasets. Our model outperformed the Conv \& KAN model, which had an accuracy of 98.75\%, for the MNIST dataset, with a test accuracy of 98.16\%. Our model obtained a test accuracy of 66.93\% for the CIFAR10 dataset, however, the accuracy of 86.29\% for the ResNet-18 model was much superior. Accuracy values of 60.26\% and 57.53\% were attained by the MLP Mixer-5 and ViT-10/4 models, respectively. Our model obtained an accuracy of 35.49\% on the CIFAR100 dataset, which is lower than the results of the ResNet-18 (59.15\%), MLP Mixer-5 (34.81\%), and ViT-10/4 (30.80\%) \cite{jeevan2022convolutional} \cite{azam2024suitability}. 

\section{Conclusion}
In summary, our research applies the KAN-Mixer architecture to show the efficiency of the Kolmogorov-Arnold Network (KAN) for vision tasks. We observed that the KAN-Mixer model achieved competitive performance, especially on the MNIST dataset, where it achieved a test accuracy of 98.16\% by conducting extensive tests on the MNIST, CIFAR10, and CIFAR100 datasets with a training batch size of 32. Even while our model performed less well than one of the state-of-the-art CNN models like ResNet-18 on the CIFAR10 and CIFAR100 datasets, it produced better results than the original MLP-Mixer. The results of this study suggest that KAN has great promise for use in vision-related tasks, and more tuning may improve its efficiency on increasingly complicated datasets.

\section{Acknowledements}
We would like to acknowledge that during our research, we were unaware of the existence of the KAN-Mixer architecture. Although we initially believed we were the first to develop this approach, we discovered that a similar architecture had been introduced just a week before the completion of our work. We extend our appreciation to the researchers who developed the KAN-Mixer and recognize their valuable contributions to advancing the field. More information about their work can be found at \url{https://github.com/engichang1467/KAN-Mixer} \cite{engichang1467}. 

\bibliographystyle{unsrtnat}
\bibliography{template}
\end{document}